\documentclass[a4paper,10pt]{article}

\usepackage{array}
\usepackage{booktabs}
\usepackage[dvipsnames]{xcolor}
\usepackage{tikz}

\newcolumntype{M}[1]{>{\centering\arraybackslash}m{#1}}

\usepackage[mlss, final, nonatbib]{cosparws_abs} 

\usepackage{subfig}

\usepackage[utf8]{inputenc} 
\usepackage[T1]{fontenc}    
\usepackage{hyperref}       
\usepackage{url}            
\usepackage{booktabs}       
\usepackage{amsfonts}       
\usepackage{nicefrac}       
\usepackage{microtype}      
\usepackage{graphicx}
\usepackage{placeins}

\usepackage{xcolor}
\newcommand{\MYhref}[3][blue]{\href{#2}{\color{#1}{#3}}}

\begin{document}

\title{Reducing Effects of Swath Gaps on Unsupervised Machine Learning Models for NASA MODIS Instruments}

\newcommand*\samethanks[1][\value{footnote}]{\footnotemark[#1]}

\author{
  Sarah Chen\thanks{Equal Contribution; Work done as researchers at Space ML} \qquad
  Esther Cao\samethanks{} \\
  \texttt{sarahc2@andrew.cmu.edu, estherca@andrew.cmu.edu} \\
  \\
  
  Anirudh Koul \qquad
  Siddha Ganju \qquad
  Satyarth Praveen  \qquad
  Meher Anand Kasam \qquad

}

\maketitle

\begin{abstract}
Due to the nature of their pathways, NASA Terra and NASA Aqua satellites capture imagery containing “swath gaps'' which are areas of no data. Swath gaps can overlap the region of interest (ROI) completely, often rendering the entire imagery unusable by Machine Learning (ML) models. This problem is further exacerbated when the ROI rarely occurs (e.g. a hurricane) and, on occurrence, is partially overlapped with a swath gap. With annotated data as supervision, a model can learn to differentiate between the area of focus and the swath gap. However, annotation is expensive and currently the vast majority of existing data is unannotated. Hence, we propose an augmentation technique that considerably removes the existence of swath gaps in order to allow CNNs to focus on the ROI, and thus successfully use data with swath gaps for training. We experiment on the UC Merced Land Use Dataset, where we add swath gaps through empty polygons (up to 20\% areas) and then apply augmentation techniques to fill the swath gaps. We compare the model trained with our augmentation techniques on the swath gap-filled data with the model trained on the original swath gap-less data and note highly augmented performance. Additionally, we perform a qualitative analysis using activation maps that visualizes the effectiveness of our trained network in not paying attention to the swath gaps. We also evaluate our results with a human baseline and show that, in certain cases, the filled swath gaps look so realistic that even a human evaluator did not distinguish between original satellite images and swath gap-filled images. Since this method is aimed at unlabeled data, it is widely generalizable and impactful for large scale unannotated datasets from various space data domains. 

\end{abstract}

\begin{figure}[ht]
  \centering
  \includegraphics[width=\textwidth]{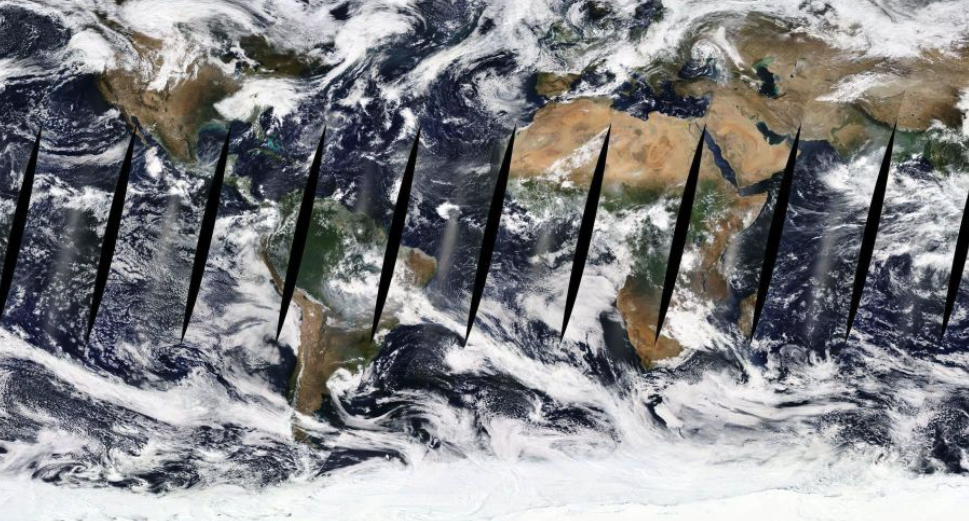}
  \caption{ \MYhref{https://worldview.earthdata.nasa.gov/}{NASA Worldview} visualizing collected satellite imagery data with black spindle-shaped swath gaps located around Earth's equator}
  \label{img:swath}
\end{figure}

\section{Introduction}

The Earth System Science is a field of study which integrates data from various interacting disciplines – e.g. land ice, oceanography, climate change – to gain a comprehensive understanding of the links between Earth’s many sub-dynamics and how they interact to form a global system. Since climate changes often occur over vast spans of time and space, the increasing availability of big data in recent years opens promising potential for this large-scale, long-term “global system” perspective in forecasting and modeling Earth-sciences phenomena. 

These advances have also aided scientists in improving the forecasting of important climate phenomena in the short-term. For example, in 1997-98, the NOAA predicted the onset of that year’s El Niño roughly 10 months in advance. To investigate the deeper question of the complex processes that contributed to the El Niño phenomena over a longer term, greater amounts of data over vast stretches of time are needed. NASA’s Earth Observing System (EOS) aims to help with this goal. 

The EOS is a group of satellites which monitor Earth and aim to collect long-term satellite imagery for this task. These satellites include the NASA Terra and Aqua satellites, which are designed specifically to study Earth’s land and water systems, respectively. 

NASA’s Moderate Resolution Imaging Spectroradiometer (MODIS) instruments are imaging sensors mounted aboard both the Terra and Aqua satellites. MODIS measures Earth’s large-scale dynamics in a wide bandwidth of wavelengths to allow nuanced measurements (e.g. cloud cover, trace gases, nutrient flow among vegetation) with moderate spatial resolution and high temporal resolution. The MODIS instrument is capable of measuring thirty six spectral bands between 0.405 and 14.38 micrometers with data acquired at three spatial resolutions: 250 m, 500 m, and 1,000 m. These satellites not only cover the entire surface of Earth every one to two days, but also work in tandem to optimize imaging under cloudy, low-light, and bright-light conditions, while minimizing optical effects such as shadows and glare. 

The Terra and Aqua instruments capture over 850 GB \cite{terra_presskit_1999} and 750 GB \cite{aqua_presskit_2002}  of data per day, accumulating over the past two decades to a total of approximately 11.5 PB of current data that is still constantly expanding. The total volume of data stored in the EOSDIS archive is estimated to be approximately 37 PB in 2020; by 2025, the total volume of data is projected to be around 250 PB\footnote{https://earthdata.nasa.gov/esds/continuous-evolution}. 

Swath gaps are empty or no data regions that occur in MODIS imagery. The first MODIS instrument, mounted aboard the Terra satellite and initially launched in 1999\footnote{https://terra.nasa.gov/}, crosses the equator at 10:30 a.m. every day. The second MODIS instrument, mounted aboard the Aqua satellite\footnote{https://aqua.nasa.gov/} and initially launched in 2002, crosses the equator at 1:30 p.m. every day. Additionally, both satellites make one full orbit every one to two days. These swath gaps exist because the MODIS satellites have a swath bandwidth of only 2330 km wide, causing consecutive orbits at the equator to miss coverage. Newer sensors, such as the Visible Infrared Imaging Radiometer Suite, that were launched in 2011 and later have no swath gaps, as they have a wider swath bandwidth of 3040 km. This bandwidth enables a roughly 15\% image overlap between consecutive equatorial orbits, covering the entire surface of Earth. 

For the MODIS satellites, these regions of missing data are primarily present in nine spindle shapes around the equator; however, they also occur at the North and South Poles during seasons of minimal sunlight. Because both satellites travel periodically between the North and South Poles, at high latitudes there is sufficient satellite data overlap to collect complete imagery of the regions. However, in regions adjacent to the equator where the circumference around the Earth at a given latitude is greater, there exist gaps between the satellites’ swath bandwidths \footnote{https://sos.noaa.gov/datasets/polar-orbiting-aqua-satellite-and-modis-swath/}. This uncollected data results is visualized as black spindles or swath gaps by NASA's Earth Observing System Data and Information System in the NASA Worldview~\footnote{https://worldview.earthdata.nasa.gov/} interface that allows browsing through all collected satellite imagery.

\subsection{Motivation}

Through data exploration experiments and input from the NASA IMPACT team, we noted that similarity search experiments that find the most similar image, supposedly based on regions of interests (ROIs) such as hurricanes or beaches, instead return images with similarly placed swath gaps. These search engines focus on swath gaps, rather than concentrating on the ROI. As such, we conclude that when notable swath gaps are present in satellite imagery, specifically used for training Earth-sciences machine learning (ML) models, these areas of missing data render the entire image unusable by unsupervised training models. Further, ML pattern recognition algorithms begin recognizing the image's swath gap as its main feature, rather than the features of its primary ROI. This is an issue because, given the nature of satellite imagery, events of interest or ROIs are already quite sporadic in regards to rare events such as tornadoes, wildfires, and volcanic eruptions, and thus every piece of data is valuable. With already limited data, the occurrence of a swath gap overlapping the ROI further reduces the available data.

\begin{figure}[ht]
  \centering
  \includegraphics[width=\textwidth]{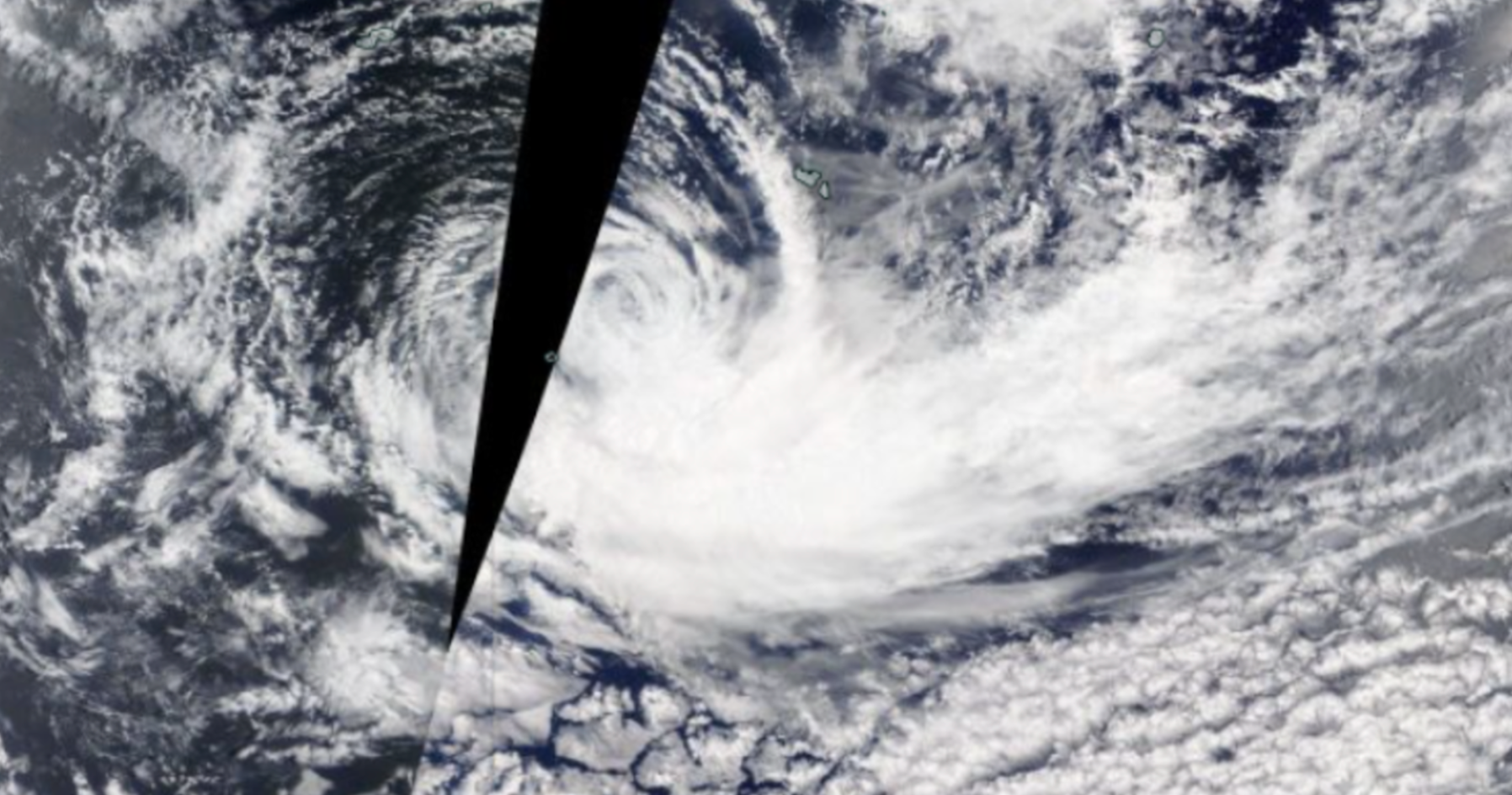}
  \caption{NASA Worldview visualizing a tiled image that contains an event of interest - a hurricane with an overlapped swath gap. We show that ML models tend to learn the features of the region of missing data, i.e. its shape and size, rather than the features of the ROI, i.e. its swirls and waves, as the identifying factors of a hurricane, and propose an augmentation technique that effectively removes the impact of the swath gap.}
  \label{img:swath}
\end{figure}

\begin{figure}[ht]
  \centering
  \includegraphics[width=\textwidth]{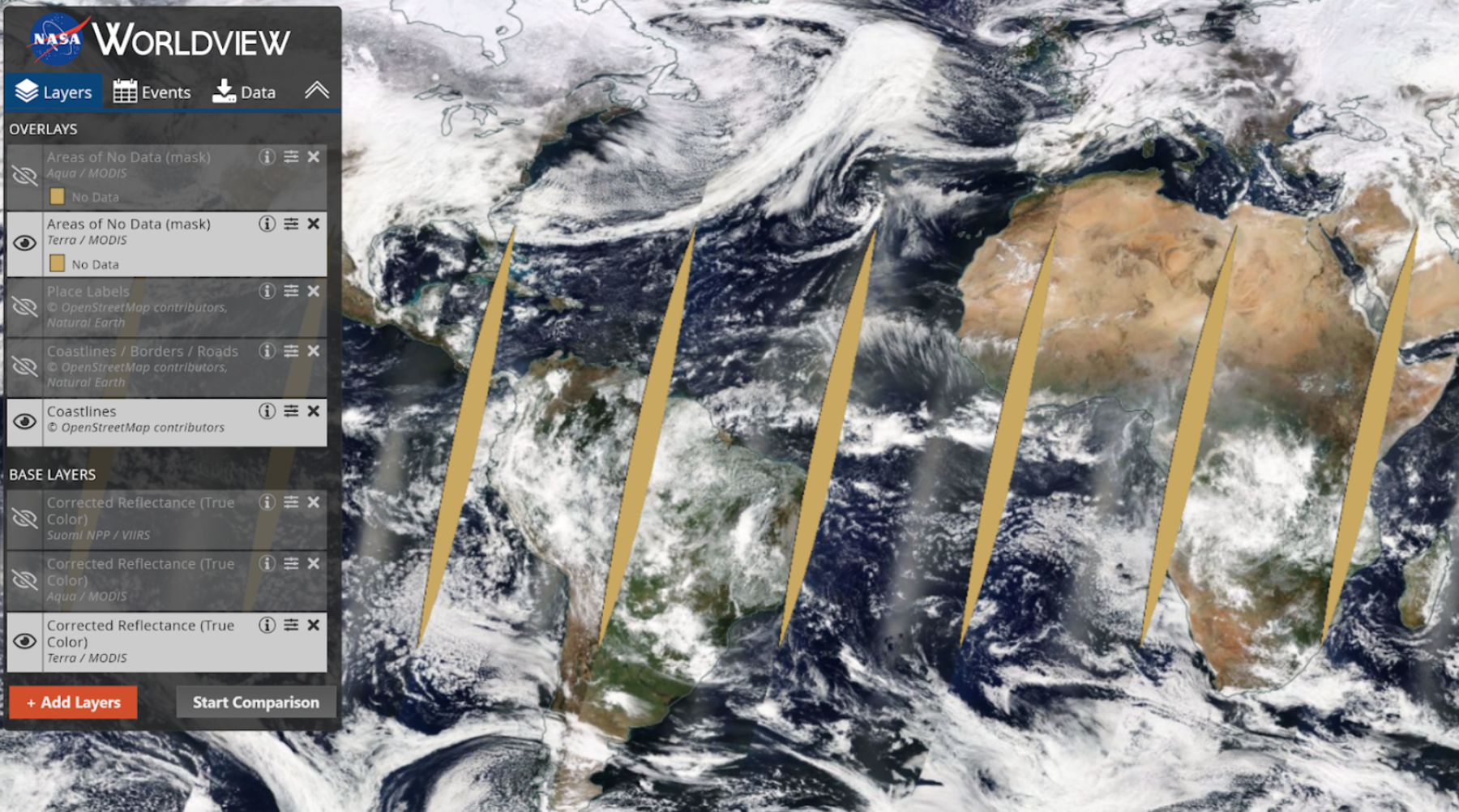}
  \caption{Worldview offers a filter ``Areas of No Data (mask)`` for Terra/MODIS and Aqua/MODIS which shows the change in the swath gap on a daily basis for each satellite.}
  \label{img:swaths_mask}
\end{figure}

Existing methods of reinforcing the ROI and helping the model differentiate between the area of focus and the swath gap include using annotated data as supervision. However, though methods are constantly evolving, annotation is still expensive and currently the vast majority of total collected data is still unannotated. Our user centered research is further motivated by inputs from the NASA IMPACT Team~\footnote{https://impact.earthdata.nasa.gov/} that show swath gaps rendering entire image sets unusable when training unsupervised ML models \cite{seeley2020kdf}.

The goal of this study is to minimize ML model attention toward swath gaps and to redirect this focus to ROIs in order to improve satellite imagery classification. We accomplish this through a unique augmentation technique that fills in the missing pixels from swath gaps and renders it almost entirely unrecognizable by the ML model. We find that it ultimately performs on par with a swath gap-less baseline. We supplement our qualitative analysis by utilizing attention maps with a human evaluator who further verifies certain cases where the filled swath gaps bear little distinction to the original satellite images.

\section{Related Works}

The work of this study builds on observations of CNNs, post-unsupervised training \cite{seeley2020kdf}. In images where the swath gaps occupy less than 25\% of the image area, outputs of the activation map convolutional layers focus more on features of the swath gaps than on those of the ROIs.

We explore the idea of filling in missing image data in order to create a seemingly more realistic image. This method is widely known as "image inpainting" and has been well explored previously \cite{criminisi2002, barnes2009}; however, the novelty of our work lies in our unique goal. We aim to fill the regions of missing data such that they are ignored by neural networks, rather than filling in swath gaps such that the entire image appears as if there were never missing data, i.e. filling the gaps in with guessed of what they were covering. The secondary technique is largely orthogonal in image inpainting methods. 

The first class of algorithms involves some degree of geometric extrapolation to realistically reproduce missing imagery. This method  extends beyond random sampling of non-missing imagery areas. Criminisi \cite{criminisi2002} first introduced the idea of combining texture synthesis algorithms and inpainting techniques in order to realistically reproduce both the three-dimensional appearance and structural integrity of a missing data region. In texture synthesis, strategies such as the Markov Random Field technique are utilized to sample each image’s non-missing pixels and reproduce a seemingly realistic texture. From classical inpainting literature, Crimininsi’s algorithm reproduces realistic structural content using mathematical models such as physical heat flow and linear extension of straight objects. Barnes \cite{barnes2009} builds upon this idea in the groundbreaking PatchMatch algorithm, which takes into account guides - such as lines, sizes, objects - specified by the user. This algorithm was further developed into image completion features and algorithms to fill in missing data that are built into Adobe Photoshop. However, for the purposes of filling in missing data from specialized ML model training imagery, both methods may introduce external artifacts into the images. For instance, in a given hurricane image, many atmospheric, fluid, and pressure dynamics that are not yet fully understood or modeled may emphasize subtle differences in the signatures of individual storms. Thus, the extrapolations which may work well for everyday objects, such as road signs and buildings, on which there exists an abundance of training data may be unacceptable when studying rarer occurrences, such as wildfires and tsunamis.

Further developments upon such techniques yield the second class of algorithms; these algorithms were primarily founded upon the use of Generative Adversarial Networks (GANs), Convolutional Neural Networks (CNNs), and Discriminatory Networks (Dnets) in order to complete a missing image region from the ground-up. For example, Sauer \cite{Sauer2016NeuralF} developed NeuralFill, a content-aware image filler which completed missing data imagery with a GAN and achieved impressive results on real-world photographs. Yu \cite{Yu_2019_ICCV} then further improved upon NeuralFill’s architecture, and later a new form of deep learning \cite{oord2016}, called “Pixel Recurrent Neural Networks” was introduced. This method  allows tractable and scalable prediction of subsequent pixels based on the surrounding, non-missing portions of an image. However, these methods of filling missing data that use generative learning run a risk of introducing artifacts or false patterns, especially while using an unbalanced or incomprehensive dataset.

The third class of methods for image filling, imputation technology, is of significant interest to this work, as it uses purely statistical methods rather than extrapolation to fill missing content. The traditional methods include filling the missing region with the average, or a random, pixel value from the background. In this work, we build upon preexisting imputation methods to develop a non deep learning-based method of filling missing image data, which “masks” the missing portions of an image using an adaptive sampling approach. This thereby eliminates the possibility of unintentionally synthesizing new artifacts that may be picked up by algorithms. 

The majority of preexisting methods rely on an underlying distribution of similar images to learn and sample from, and this distribution sample is largely unavailable in unsupervised settings. Additionally, we refrain from attempting to fill the swath gaps with synthetic data, e.g. filling the eye of a hurricane with similar pixel values, as this method could potentially introduce external artifacts and, thus, skew the ML model’s detection of detailed information. This procedural error would leave possibilities of filling the empty gaps with alternative data that could be recognized by pattern-finding algorithms, compromising the integrity of the images. 

\section{Experiments}

Here we describe our experimental setup, our main motivational experiment, and the iterative augmentation techniques that improved the ML models’ emphases on the ROI rather than the swath gaps.

\subsection{Data}

\begin{figure}[ht]
  \centering
  \includegraphics[width=60mm]{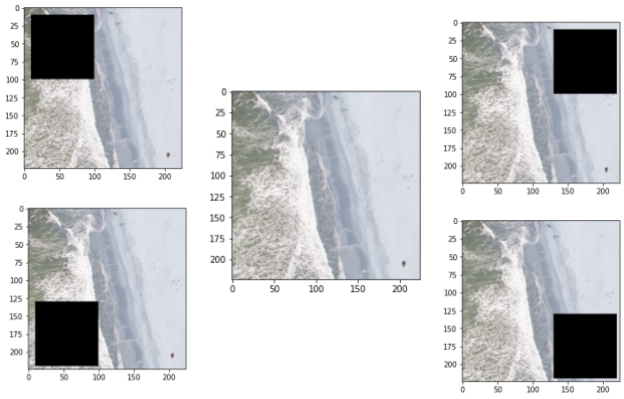}
  \caption{For each image in the dataset, we add five ‘copies’ of the same image to the dataset, each containing a different swath gap position: no swath gap, upper left, upper right, lower left, and lower right. Squares of missing data are equally sized for standardization. }
  \label{img:beach_swath_locations}
\end{figure}

Though the MODIS satellite data includes swath gaps, it is unusable for training and testing due to the fact that it does not provide a golden set, i.e. an identical dataset without swath gaps. Additionally, the raw MODIS data poses a challenge of labels - a vast majority of the satellite imagery is unlabeled - which would make assessing our results difficult. As such, we decide to use a labeled dataset to facilitate our process of evaluating the success and accuracy of our augmentation techniques. Further, we simulate our own swath gaps and build conditions comparable to those visualized in NASA Worldview by generating a new dataset in which a portion of each image contains a swath gap. Thus, we utilize the UC Merced Land Use Dataset~\cite{yang2010bselc} -- a labeled dataset with images from various land-related categories -- and we simulate the swath gaps as black boxes of no data, inlaid in one of the four corners (\ref{img:beach_swath_locations}). Our dataset is a subset of the UC Merced Land Use dataset with five randomly selected images from each of seven distinct classes, i.e. `airplane`, `beach`, `forest`, `harbor`, `freeway`, `river`, `storagetanks`. This modest subset of the entire UC Merced Land Use dataset allows complete utilization of the limited compute resource available through freemium Google Colaboratory. 

For each image in our dataset, a swath gap is generated in one of the following positions: left-lower, left-upper, right-lower, right-upper, or absent swath gap. These positionings are employed for autoencoder evaluation purposes: we can quantify how much the model emphasizes swath gaps by counting the number of images with swath gaps in the same corner, post-similarity search. Additionally, swath gaps are represented as null-value filled square shapes that comprise 20\% of each image's area. This replicates real-world conditions of MODIS satellites where only swath gaps that occupy less than 25\% of missing data are considered usable in satellite image categorization. For the autoencoder model described later in the paper, data were partitioned into 90\% training and 10\% validation data.

\paragraph{Experimental Setup:}
We utilize the ResNet50 pre-trained model \cite{he2015}, and fine tune this model on a subset of the UC Merced Land Use dataset. While not specifically trained on Earth satellite imagery, the ResNet50 provides an accessible and time-efficient assessment of how a supervised model may train on our data. We train our models using the NVIDIA Tesla K80 GPU provided by Google Colaboratory. We also release all the code and models used for our experiments on Github \footnote{https://github.com/spaceml-org/Missing-Pixel-Filler}.

\paragraph{Analysis with Activation Maps}
We describe a simple experiment which displays how the ML model responds firstly to swath gaps and then to ROIs. We model the problem using a similarity search with a query image that contains a swath gap. The ground-truth system for comparison is a ResNet50, fine-tuned with a golden dataset, i.e. one with no swath gaps. We note through visual inspection of the results that the trained ML model (1) is unable to satisfactorily retrieve the correct class category and (2) recognizes the location of the swath gaps as the primary feature of the query image. Thus, the model retrieves images with similarly located swath gaps rather than images with similar ROI features.

We further verify our initial conclusion with activation maps. Activation maps act as pattern-detectors that are generated during training and highlight the aspects of an image that are most heavily weighted by the respective ML model. In this study, activation maps are utilized for preliminary visual evaluation. We use the pre-trained ResNet50 model to run inference with swath gap-containing images. Then, we extract the activation maps of the CNN’s training and obtain the maps shown in Figure \ref{tbl:table_of_figures}. By examining activation maps trained on image data that contain swath gaps, we observe that the region of missing data is highlighted in the activation map. This indicates that the model is focusing primarily on the swath gap. The activation maps serve as a visual inspection to understand and interpret the quality of various swath gap-filling augmentation techniques. We further note that our swath gaps are completely and uniformly distributed, i.e. all pixels are $(225,225,225)$, and they appear in consistent locations across all class categories; thus hypothetically, the ML model should learn to ignore them, as the swath gaps do not provide any unique information or differentiating factor between different image class categories. However, we acknowledge that they create an edge wherever they occur. 

\begin{table}
    \centering
    \begin{tabular}{cM{20mm}M{20mm}M{20mm}M{20mm}}
       \toprule
        & Original Image & Original Image - Activation Map & Image with Swath Gap & Image with Swath Gap - Activation Map \\
        \midrule
        Airplane & \includegraphics[width=20mm]{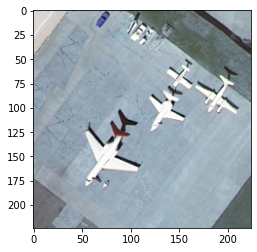} & \includegraphics[width=20mm]{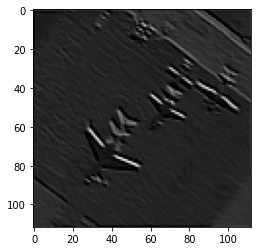} & \includegraphics[width=20mm]{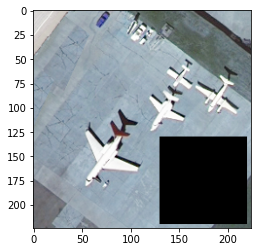} & \includegraphics[width=20mm]{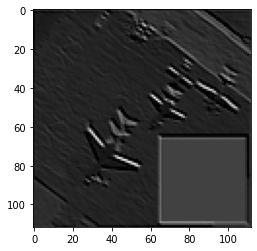} \\
        Beach & \includegraphics[width=20mm]{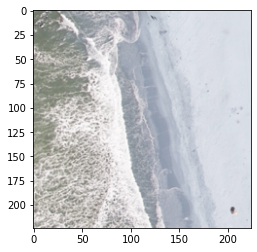} & \includegraphics[width=20mm]{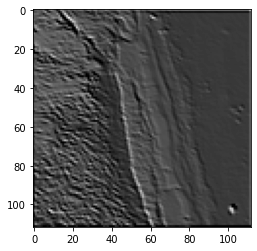} & \includegraphics[width=20mm]{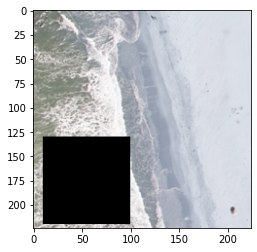} & \includegraphics[width=20mm]{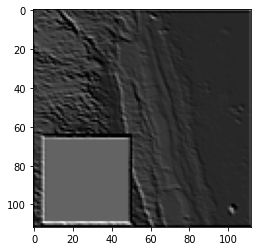} \\
        Forest & \includegraphics[width=20mm]{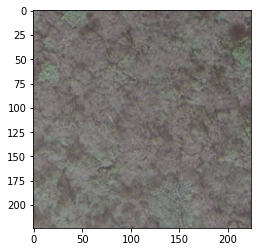} & \includegraphics[width=20mm]{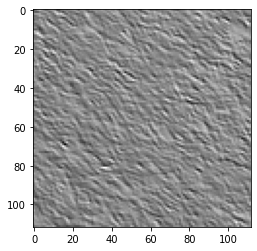} & \includegraphics[width=20mm]{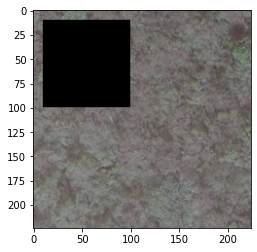} & \includegraphics[width=20mm]{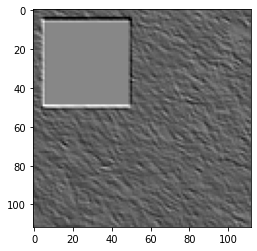} \\
        Freeway & \includegraphics[width=20mm]{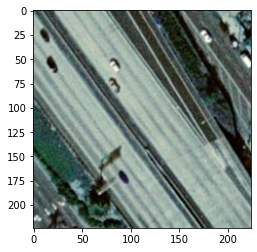} & \includegraphics[width=20mm]{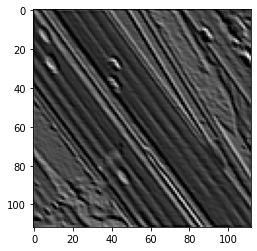} & \includegraphics[width=20mm]{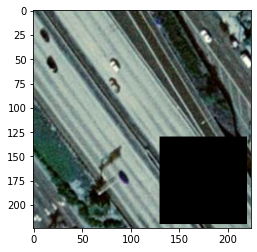} & \includegraphics[width=20mm]{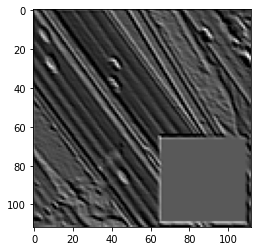} \\
        Harbor & \includegraphics[width=20mm]{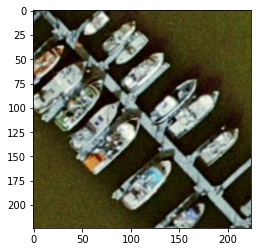} & \includegraphics[width=20mm]{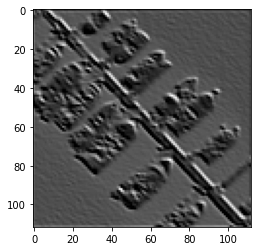} & \includegraphics[width=20mm]{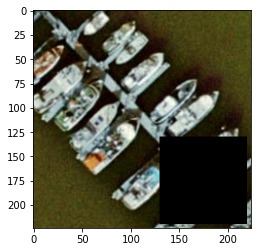} & \includegraphics[width=20mm]{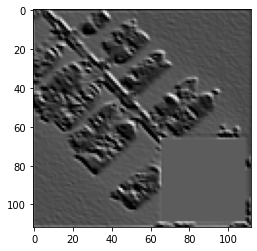} \\
        River & \includegraphics[width=20mm]{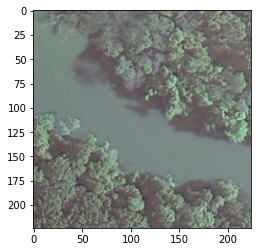} & \includegraphics[width=20mm]{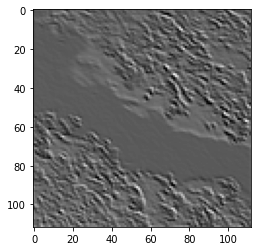} & \includegraphics[width=20mm]{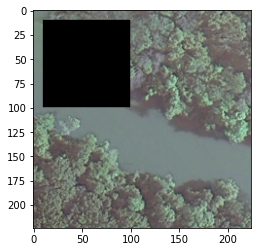} & \includegraphics[width=20mm]{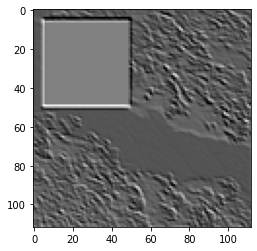} \\
        Storage Tanks & \includegraphics[width=20mm]{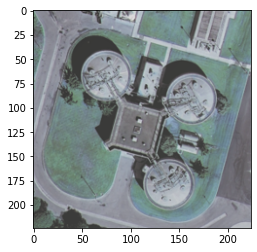} & \includegraphics[width=20mm]{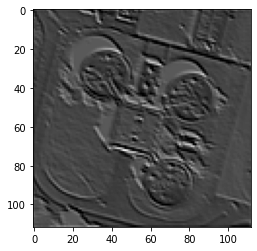} & \includegraphics[width=20mm]{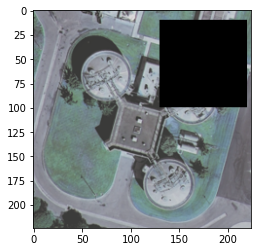} & \includegraphics[width=20mm]{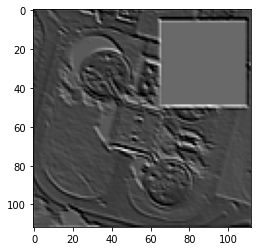} \\
        \bottomrule
    \end{tabular}
    \caption{The swath gaps stand out in the activation maps; this implies that the model is focusing on these swath gaps, causing the model to learn the location during training. In an unsupervised setting, the model could potentially learn that images belong to the same category if they contain similarly located swath gaps, rather than if the images contain the class category’s ROI such as an `airplane` or `beach`. We build on iterative augmentation techniques that `deactivate,` or attempt to reduce the impact of, the swath gaps.}
    \label{tbl:table_of_figures}
\end{table}

\subsection{Motivation: Evaluating a Naive Model's response to Swath Gaps}
We conduct similarity searches on images with and without swath gaps. The ground truth is a similarity search performed on an image without swath gaps from various class categories. Note, all the retrieved images contain the same class category with high accuracy. We compare these results to a similar search where the query images have swath gaps. Note, the class category of the retrieved images are different from the class category of the query image, and locations of the swath gaps are identical.

\begin{figure}[ht]
  \centering
  \includegraphics[width=\textwidth]{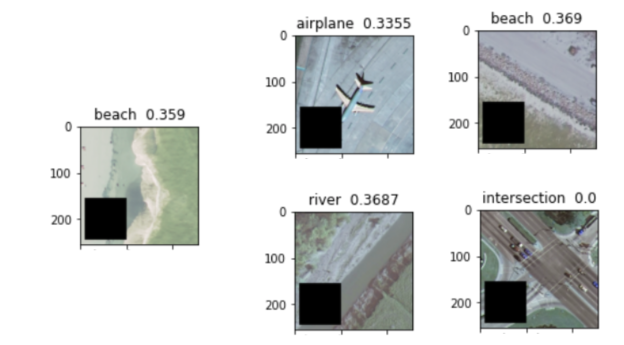}
  \caption{With the presence of the missing data, the images returned from similarity searches indicate that the algorithm recognizes the location of the swath gaps as the primary feature of these images, rather than the recognizing the identifying features of the ROI.}
  \label{img:bad_sim_search_results}
\end{figure}

\subsection{Augmentation Techniques to Fill the Swath Gaps}
With our prior conclusion that a ML model selectively focuses on swath gaps in an unsupervised setting, we iteratively experiment with various ways to deactivate, or reduce the impact of, the swath gap. Hence, we decide that the best way to approach our problem is to fill the empty data with a featureless pattern that results in no pattern recognition at all surrounding the swath gap (Figure \ref{tbl:table_of_figures}.

\paragraph{Policy 1: Random RGB} We replace each pixel in the swath gap with a random pixel value from a uniform distribution. This attempts to randomly fill the entire swath gap such that there is no inherent pattern the model could potentially recognize. Though no pattern is visibly recognized inside the swath gap, the swath gap itself is recognizable: the edges of the random RGB-filled swath gap still stand out, and thus preemptively we conclude that this policy is insufficient in obscuring the swath gap. 

\paragraph{Policy 2: Pixel RGB} Instead of replacing each missing pixel in the swath gap with a random RGB value as before, we propose a new solution method to replace the pixels with random RGB values already present within the image, excluding the swath gap portion. The pixels to replace the swath gap are sampled from the image’s histogram, aligning with its pixel intensity values. No visible pattern is recognized within the swath gaps; however, the swath gaps can still be distinctly recognized by its obvious borders. 

\paragraph{Policy 3: Neighbor RGB}
Finally, we opt to replace the swath gap pixels with the neighboring pixel values - i.e. random pixels are picked from within a certain radius of the swath gap. The radius value changes depending on how large the swath gap is and how distant the target pixel of the swath gap is from its border. Thus, a random value is drawn from the given radius, for a maximum of three times. If no non-null value has been chosen, then the radius is expanded. This process continues until the entire swath gap is filled. Visualizing this dynamic fill method, we see that the swath gap has neither a perceivable pattern within the region nor a clearly distinguishable edge that can be recognized by the model. 

\begin{table}
    \centering
    \begin{tabular}{M{20mm}M{20mm}M{20mm}M{20mm}}
       \toprule
        \includegraphics[width=20mm]{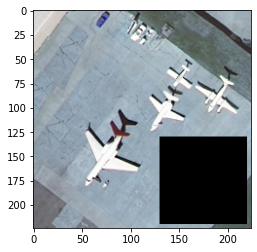} & \includegraphics[width=20mm]{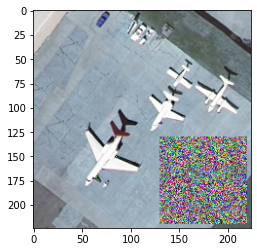} & \includegraphics[width=20mm]{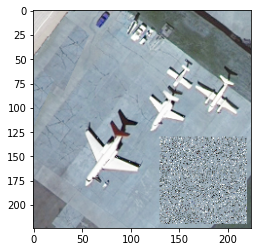} & \includegraphics[width=20mm]{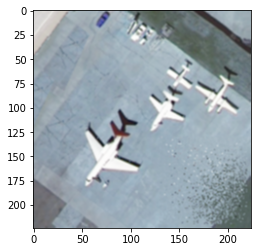} \\
        \includegraphics[width=20mm]{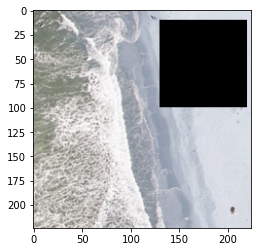} & \includegraphics[width=20mm]{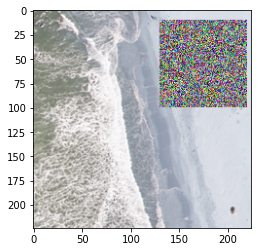} & \includegraphics[width=20mm]{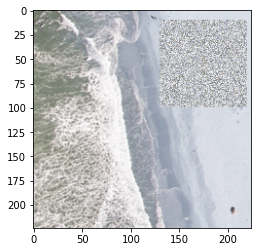} & \includegraphics[width=20mm]{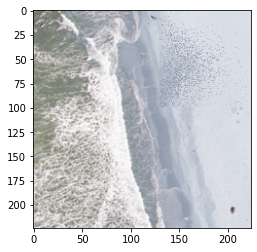} \\
        \bottomrule
    \end{tabular}
    \caption{Applying the Swath Filling policies. From left to right: Swath Gap, Random RGB-filled, Pixel RGB-filled, Neighbor RGB-filled. \textbf{Random RGB} fills the swath gap with randomly selected RGB pixel values from a uniform distribution. Visually, there is no recognizable pattern inside the swath gap. However, the swath gap itself remains readily distinguishable. \textbf{Pixel RGB} fills the swath gap with randomly selected pixel values from the image. Visually, again, there is no recognizable pattern inside the swath gap, and the RGB values of the gap align more closely with the RGB values of the image. However, the swath gap itself is still distinguishable. \textbf{Neighbor RGB} dynamically fills the swath gap with random pixel values from the image, following a selection criteria detailed in this paper. There is neither a recognizable pattern inside the swath gap nor an obvious border that causes the region of filled-in data to stand out.}
    \label{tbl:table_of_figures}
\end{table}

\subsection{Activation Maps and Similarity Searches}

In the simulated dataset, swath gaps were filled with each of the three strategies. Then, each method's effectiveness in obscuring the swath gaps was assessed using activation maps and similarity searches.

\paragraph{A. Similarity Search}
Similarity searches are conducted using autoencoders. Autoencoders are a type of self-supervised model commonly used for their simplicity, adaptability, and unsupervised nature. 

We use an autoencoder to perform similarity searches - given an input image, we ask the autoencoder to return the four most similar images from the dataset. Separate autoencoder models with each augmentation policy are trained.

The ground truth is the original image, absent of any swath gap. The search performed with high accuracy and returns four images of real beaches. However, when the search was re-conducted on another beach image, with simulated swath gaps in the bottom left corner, results differed vastly. 

When satellite imagery contains missing data, the similarity search returns images based on the swath gaps', rather than the ROIs’, features. For instance, query images with upper-left swath gaps tended to also return similarity search images with upper-left swath gaps, regardless of their respective ROIs. Reiterations of similarity searches on various query images with swath gaps confirmed that the computer algorithm was often recognizing swath gap locations as the primary feature of missing data images. This confirmed the aforementioned flaw - according to a ML model, swath gaps identify the image and thus disproportionately capture the attention of computer vision models and divert bandwidth from important features of the image.

Testing was then conducted on the three filling methods by querying images filled with our methods into the autoencoder. These results greatly improved as methodology increased from method one (Random RGB) to two (Pixel RGB) to three (Neighbor RGB), with successively more images categorized correctly per fill method. Repetition of this experiment indicated that filling method three, Neighbor RGB, was the most efficient, with consistently three or four out of the four "most similar" images being categorized correctly.

\begin{table}
    \centering
    \begin{tabular}{cM{20mm}M{100mm}M{20mm}M{20mm}}
       \toprule
        & Successes (out of 4) & Query Image (L) with its 4 Similarity Matches (R) \\
        \midrule
        Original & 4 & \includegraphics[width=100mm]{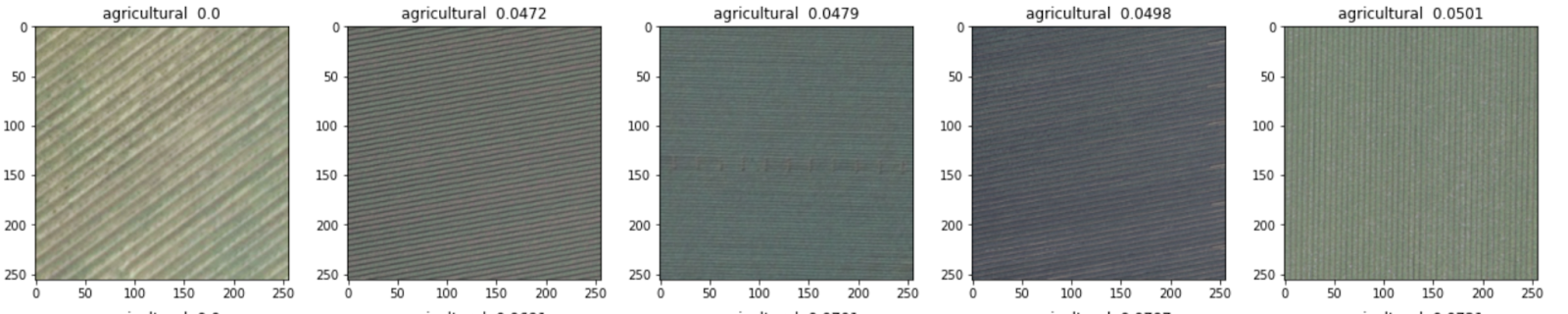} \\
        No Fill & 0 & \includegraphics[width=100mm]{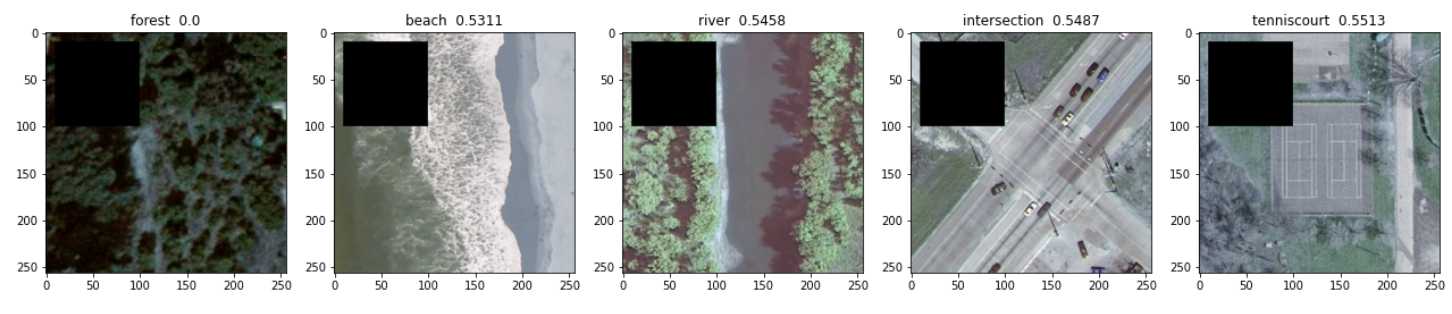} \\
        Random Fill & 1 & \includegraphics[width=100mm]{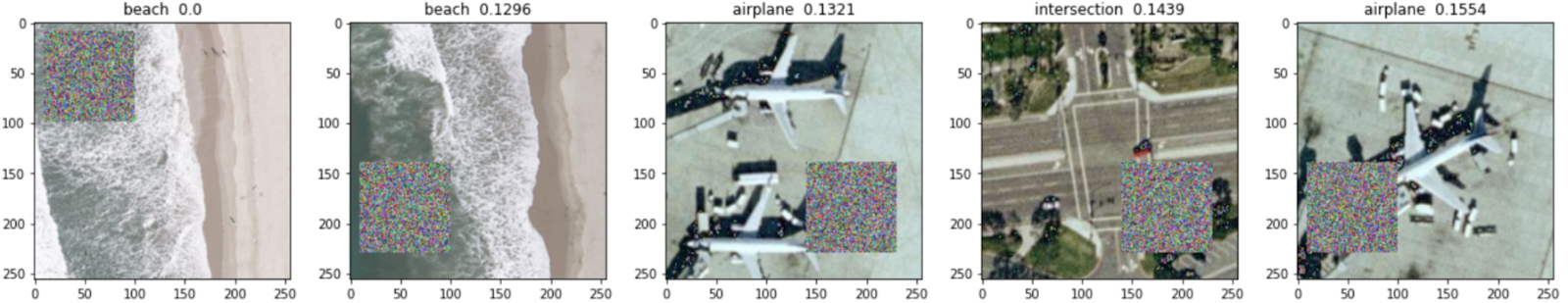} \\
        Pixel Fill & 2 & \includegraphics[width=100mm]{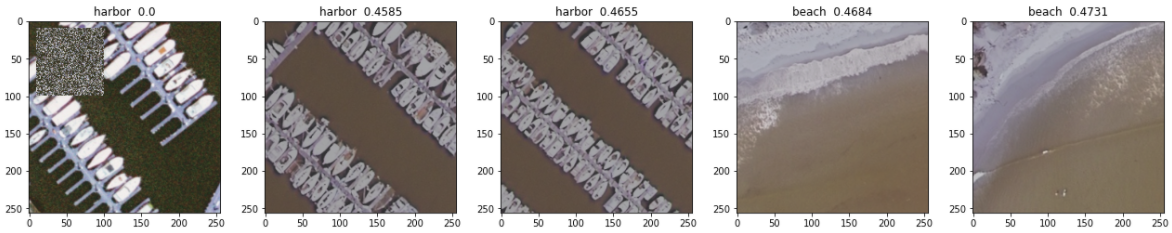} \\
        Neighbor Fill & 4 & \includegraphics[width=100mm]{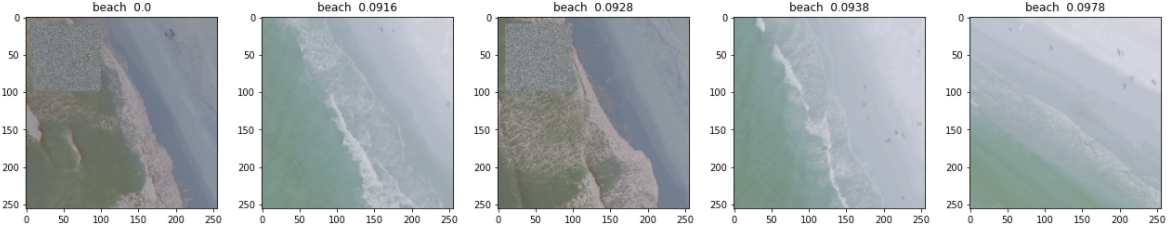} \\
        \bottomrule
    \end{tabular}
    \caption{Trained convolutional autoencoder outputs. Query image (leftmost column) and its corresponding most-similar four images. Filling strategy changes row wise: no fill, Random RGB, Pixel RGB, Neighbor RGB. Random RGB fill strategy results show that the autoencoder focuses on swath gap positions. Neighbor RGB fill strategy results show that the autoencoder ignores the swath gap and concentrates on the ROI.}
    \label{tbl:table_of_figures}
\end{table}

Therefore results emphasize that swath gaps, despite seeming small in size, are significant problems that need to be addressed in order for computer vision models to properly classify satellite imagery. 

\subsection{Qualitative Analysis: Visualizing Augmentation Strategies with Activation Maps}
In the figures below, images on the left are the original while images on the right are their corresponding activation maps. In the swath gap image below, the gaps stand extremely out in the activation map. It is evident that the map is paying way too much attention to these regions of missing data, causing the model to learn the location of the swath gaps as the images’ identifying feature during training. This means that the model assumes that images belong to the same category if they contain similarly placed swath gaps, rather than if the images have similar features that identify what the ROI and satellite imagery actually is. 

For the first filling method, although the activation maps were focused less on the random RGB inside the swath gap, they were still very focused on the edges. The second filling method has a similar problem, although less pronounced. However, in the third filling method, the swath gap is nearly indistinguishable on the activation map. This qualitatively indicates that the unsupervised model is paying few attention, if any, to the swath gap under the third filling method.

\begin{table}
    \centering
    \begin{tabular}{cM{20mm}M{20mm}M{20mm}M{20mm}}
       \toprule
        & Swath Gap Image & Method 1: Random RGB & Method 2: Pixel RGB & Method 3: Neighbor RGB \\
        \midrule
        Original Airplanes & \includegraphics[width=20mm]{a11} & \includegraphics[width=20mm]{a12} & \includegraphics[width=20mm]{a13} & \includegraphics[width=20mm]{a14} \\
        Activation Maps of Airplanes & \includegraphics[width=20mm]{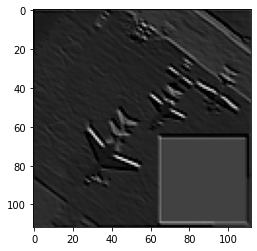} & \includegraphics[width=20mm]{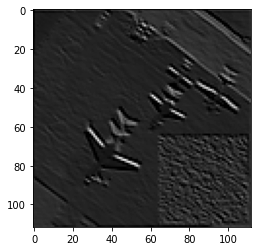} & \includegraphics[width=20mm]{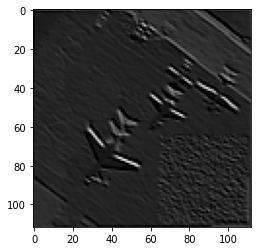} & \includegraphics[width=20mm]{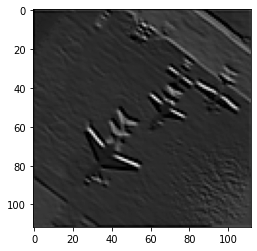} \\
        Original Beach & \includegraphics[width=20mm]{b11} & \includegraphics[width=20mm]{b12} & \includegraphics[width=20mm]{b13} & \includegraphics[width=20mm]{b14} \\
        Activation Maps of Beaches & \includegraphics[width=20mm]{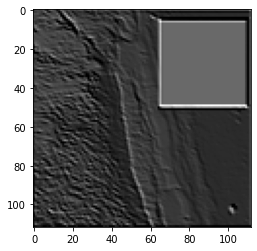} & \includegraphics[width=20mm]{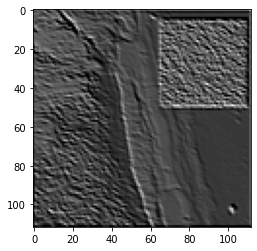} & \includegraphics[width=20mm]{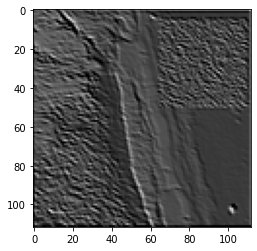} & \includegraphics[width=20mm]{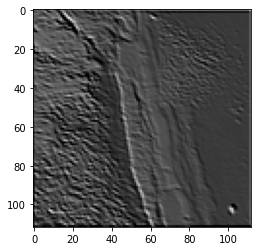} \\
        \bottomrule
    \end{tabular}
    \caption{Trained convolutional autoencoder outputs. Query image (leftmost column) and its corresponding most-similar four images. Filling strategy changes row wise: no fill, Random RGB, Pixel RGB, Neighbor RGB. Random RGB fill strategy results show that the autoencoder focuses on swath gap positions. Neighbor RGB fill strategy results show that the autoencoder ignores the swath gap and concentrates on the ROI.}
    \label{tbl:table_of_figures}
\end{table}

\subsection{Conclusion}
Based on results from activation map observations and autoencoder similarity searches, the Neighbor RGB strategy is the best filling method. This strategy fills simulated swath gaps with a dynamic algorithm that successfully obscures regions of missing data.

In supervised machine learning, input images are often already labeled. Thus, supervised networks already have patterns and features associated with each respective label and can classify new images based on this bank of knowledge. Thus, swath gaps pose less significant of a problem there. However, this study is crucial in regards to unsupervised ML models. In unsupervised machine learning, training images do not have labels. The neural network has no metric of comparison or accuracy detection, so the network retains the most prominent feature of the image as its identifying pattern in order to “learn” how to classify an image. Since this "most prominent feature" is often recognized as the swath gap, i.e. area of missing data, and not the ROI, missing data pose a large problem. Thus, the methodology of this study is critical in obscuring swath gaps with unsupervised ML. Hence, the results not only apply to the simulated dataset of real Worldview images, but also extend to other domains of work regarding unsupervised ML models.

Though our findings result in a best filling strategy, this study has weak points to be addressed in future work: our dataset was a fairly modest replica of Worldview data, and more quantifiable evidence would have strengthened results. Additionally, swath gap sizes in this study were restricted to consistent rectangles that comprised exactly 20\% of the image.

On the other hand, this study also includes strong points: by not using a pre-created generative model to fill the swath gaps, the possibility that models could have generated a pattern inconsistent with the original image was avoided. This would have compromised the integrity of the image and could have potentially created new patterns that distracted CNN attention from the real ROI. Additionally, using an autoencoder similarity search to measure our success was beneficial, as this process mimics unsupervised CNN pattern recognition.

\section{Future Work}
Due to the lack of large scale compute resources and the limited tier of freemium Google Colaboratory GPUs, utilizing our findings and replicating similar results on a large scale MODIS imagery data has been left for future work. Additionally, in conjunction with the NASA IMPACT team, we plan on pursuing a deeper analysis of the MODIS data, building on the qualitative and quantitative results and improving the Machine Learning Technology Readiness Level \cite{alex2021technology} of our system. We acknowledge that the evidence in support of our claims need much more strengthening; therefore, we are currently in the process of testing our algorithms on a larger dataset with automated training and evaluations utilizing the NASA GIBS Worldview Similarity Search \cite{venguswamy2021} to gather more results in support of our results. 

\section{Acknowledgement}
We thank the SpaceML~\cite{koul2021spaceml} team for helpful discussions, guidance, and dedicated mentorship, and we thank Rahul Ramachandran and his team from the larger NASA IMPACT team for their domain expertise and continued guidance and the initial research by the Frontier Development Lab~\cite{Ganju2020FDL} Knowledge Discovery Framework ~\cite{seeley2020kdf}.

\bibliographystyle{IEEEtran}
\bibliography{references}

\end{document}